\documentclass[letterpaper]{article} 
\usepackage{aaai24} 
\usepackage{times}  
\usepackage{helvet}  
\usepackage{courier}  
\usepackage[hyphens]{url}  
\usepackage{graphicx} 
\usepackage{natbib}  
\usepackage{caption} 
\usepackage{algorithm}
\usepackage{algorithmic}

\usepackage{newfloat}
\usepackage{listings}
\DeclareCaptionStyle{ruled}{labelfont=normalfont,labelsep=colon,strut=off} 
\floatstyle{ruled}
\newfloat{listing}{tb}{lst}{}
\floatname{listing}{Listing}
\usepackage{inconsolata}

\usepackage{amsmath}
\usepackage{amssymb}
\usepackage{booktabs}

\usepackage{soul}
\usepackage{xspace}
\usepackage[dvipsnames]{xcolor}
\usepackage{array}
\usepackage{multirow}
\usepackage{subcaption}

\usepackage[capitalize]{cleveref}
\crefname{section}{Sec.}{Secs.}
\Crefname{section}{Section}{Sections}
\Crefname{table}{Table}{Tables}
\crefname{table}{Tab.}{Tabs.}

\makeatletter
\DeclareRobustCommand\onedot{\futurelet\@let@token\@onedot}
\def\@onedot{\ifx\@let@token.\else.\null\fi\xspace}
\def\eg{\emph{e.g}\onedot} 
\def\ie{\emph{i.e}\onedot} 
 
 \def\vs{\emph{vs}\onedot}
\def\wrt{w.r.t\onedot}

\makeatother

\def\approach{DEED}
\def\ours{DEED~}
\def\oursnospace{DEED}
\def\th{$\tau$~}
\def\thnospace{$\tau$}

\title{\approach: Dynamic Early Exit on Decoder \\for Accelerating Encoder-Decoder Transformer Models}
\author{Peng Tang$^{\dagger}$ \quad Pengkai Zhu$^{\dagger}$ \quad Tian Li$^{\ddagger}$\thanks{Work conducted during an internship at Amazon.} \quad Srikar Appalaraju$^{\dagger}$ \\ Vijay Mahadevan$^{\dagger}$ \quad R. Manmatha$^{\dagger}$\\
$^{\dagger}$AWS AI Labs \quad $^{\ddagger}$University of California San Diego\\
{\tt\small \{tangpen, zhpengka, srikara, vmahad, manmatha\}@amazon.com \quad tianli@ucsd.edu}
}
\usepackage{bibentry}

\begin{document}

\maketitle

\begin{abstract}
   Encoder-decoder transformer models have achieved great success on various vision-language (VL) tasks, but they suffer from high inference latency. Typically, the decoder takes up most of the latency because of the auto-regressive decoding. To accelerate the inference, we propose an approach of performing Dynamic Early Exit on Decoder (DEED). We build a multi-exit encoder-decoder transformer model which is trained with deep supervision so that each of its decoder layers is capable of generating plausible predictions. In addition, we leverage simple yet practical techniques, including shared generation head and adaptation modules, to keep accuracy when exiting at shallow decoder layers.
   Based on the multi-exit model, we perform step-level dynamic early exit during inference, where the model may decide to use fewer decoder layers based on its confidence of the current layer at each individual decoding step. Considering different number of decoder layers may be used at different decoding steps, we compute deeper-layer decoder features of previous decoding steps just-in-time, which ensures the features from different decoding steps are semantically aligned. We evaluate our approach with two state-of-the-art encoder-decoder transformer models on various VL tasks.
   We show our approach can reduce overall inference latency by 30\%-60\% with comparable or even higher accuracy compared to baselines.
\end{abstract}

\section{Introduction}
\label{sec:intro}

\begin{figure}[t]
    \centering
    \includegraphics[width=0.93\linewidth]{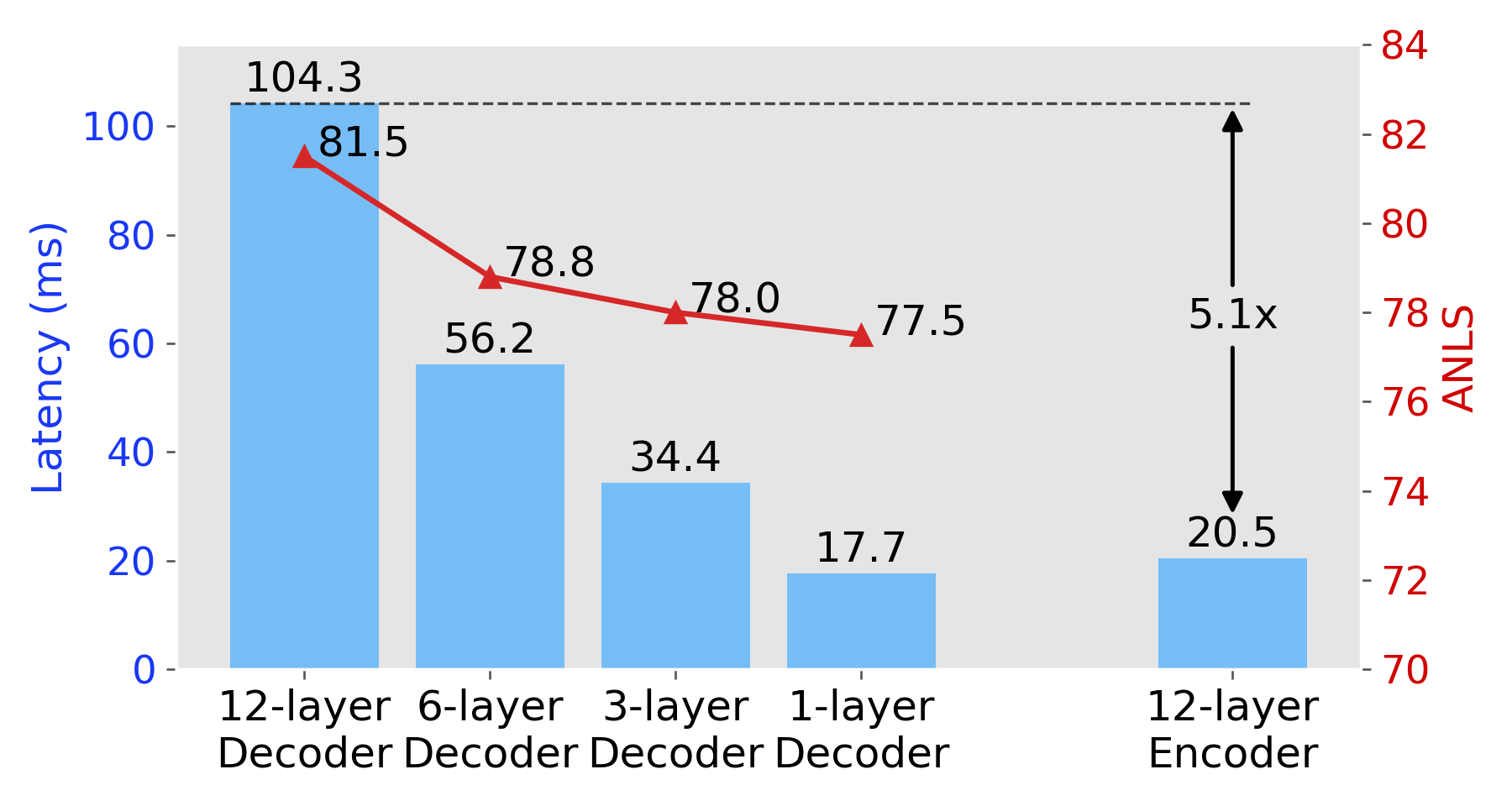}
    \caption{Inference latency and accuracy (ANLS \cite{mathew2021docvqa}) of LaTr++\cite{biten2022latr} using different number of decoder layers on the DocVQA validation set. The decoder takes most of the inference time compared to the encoder (104.3 ms \vs 20.5 ms). In addition, even using one decoder layer can achieve decent accuracy (77.5\% \vs 81.5\%), implying that most examples do not need all decoder layers during inference.}
    \label{fig:lowLayerHighAcc}
\end{figure}

Vision-Language (VL) tasks, \eg, Visual Question Answering (VQA) \cite{balanced_vqa_v2,biten2019scene,mathew2021docvqa,mishra2019ocr,singh2019towards} and referring expression comprehension \cite{mao2016generation,yu2016modeling}, have drawn increasing attention in recent years. These tasks involve reasoning about image and texts at the same time. Among many successful models \cite{appalaraju2021docformer,chen2020uniter,wang2022ofa,zhang2021vinvl,biten2022latr,lu2022unified,chen2022pali,zhou2020unified} tackling these tasks, encoder-decoder transformer models \cite{wang2022ofa,biten2022latr,chen2022pali} usually show the best accuracy thanks to the strong generative ability of the decoder.

Nevertheless, encoder-decoder models rely on the auto-regressive decoding to bring its ability into full play at inference. With auto-regressive decoding, each output token is generated conditioned on previous tokens. Therefore, it has to generate tokens one after another, and repeat the feed-forward in each layer as many times. This mechanism leads to high inference latency in the decoder, and makes the decoder take up most of the total inference latency, as shown in Figure \ref{fig:lowLayerHighAcc}. Interestingly, even using only one decoder layer, an encoder-decoder model can still get decent prediction accuracy (see Figure \ref{fig:lowLayerHighAcc}), which means samples got correct by the one decoder layer do not need the excessive computations in the deeper decoder layers.

Inspired by these facts, we propose an approach to dynamically allocate adequate amount of computation at a particular decoding step in order to speed-up inference without sacrificing accuracy. Specifically, we build Dynamic Early Exit on Decoder (\approach), a multi-exit model with an early exit strategy to let the model decide whether or not to exit at a specific decoder layer at each decoding step dynamically.
Following existing work \cite{xin2020deebert,Liu_Meng_Zhou_Chen_Xu_2021,liu2020fastbert,zhang-etal-2022-pcee,geng2021romebert,xin2021berxit,zhou2020bert}, we employ confidence-based dynamic early exit where the decoder may decide to exit when it is confident about its prediction. Unlike encoder acceleration, the dynamic early exit for auto-regressive decoder is more challenging. The challenge is two-fold:
\begin{itemize}
    \item Multi-exit model, \ie, a model that can exit / make prediction at each layer. To get accurate predictions out of dynamic early exit, we must build and train a strong multi-exit encoder-decoder model, where each of the decoder layers have strong generative ability.
    \item  Semantic misalignment at inference. Tokens can be generated at different decoder layers at different decoding steps. But the auto-regressive decoding requires $n$-layer features from all the previous steps if the current step is inferring at layer $n$. They won't be available if the previous decoding steps exit at shallower layers. This semantic misalignment between different layers imposes difficulties when applying naive early exit strategy, leading to degraded accuracy.
\end{itemize}

Previous approaches address the first challenge by using different prediction heads after each transformer layer \cite{schwartz2020right,xin2020deebert,geng2021romebert,xin2021berxit}.
In contrast, we build our multi-exit model by sharing the generation head among different decoder layers and training with deep supervision.
The generation head generates the output sequence prediction, \eg, the answer text for visual question answering \cite{biten2022latr,chen2022pali,alayrac2022flamingo,lu2022unified} or box coordinates for referring expression comprehension \cite{wang2022ofa,lu2022unified}.
In addition, we insert an adaptation module between decoder layers and the generation head.
This design helps to strengthen the generative ability of shallow decoder layers by sharing the common generation knowledge among different decoder layers.
Moreover, to maintain the generative ability when exiting at the final layer, we proposes a loss function that emphasizes the learning of the final layer.  
These simple yet effective techniques help to improve the accuracy of shallow decoder layers without sacrificing the accuracy at the final decoder layer.

To address the second challenge, we propose a novel algorithm that dynamically computes required deeper-layer features for previous decoding steps just-in-time. This algorithm effectively resolves the semantic misalignment among different layers at different generation steps. In contrast, the existing work, Depth-Adaptive Transformer (DAT)~\cite{elbayad2019depth}, which uses the shallow-layer features as the deeper-layer features directly for later decoding steps, failed to mitigate such semantic misalignment and thus substantially undermine the generative ability of the model.  

Our contributions are summarized as follows:
\begin{itemize}
    \item We propose \oursnospace, a multi-exit model with step-level dynamic early exit on decoder to speed-up inference without sacrificing accuracy for encoder-decoder transformer models.
    \item We apply our approach to two state-of-the-art encoder-decoder transformer models and evaluate on various VL datasets. Our approach is able to reduce 30\%-60\% overall inference latency with comparable or even higher accuracy compared to baseline models and other dynamic early exit approaches.
    \item Our approach provides a trade-off between accuracy and latency by using a variable confidence threshold.
\end{itemize}

\section{Related Work}
\label{sec:related}

\noindent{\bf Encoder-Decoder Models for Vision-Language Tasks}
Encoder-decoder transformer models have pushed the edge for Vision-Language (VL) tasks recently \cite{alayrac2022flamingo,wang2022ofa,biten2022latr,lu2022unified,chen2022pali} because of strong representation ability of encoder and generative ability of decoder.
For example, Flamingo \cite{alayrac2022flamingo} uses a vision encoder to encode input images and a text decoder to generate text predictions for various VL tasks.
LaTr \cite{biten2022latr} utilizes the sequence generation ability in decoder and layout in multi-modality learning and achieves state-of-the-art accuracy on text-based VQA tasks.
OFA \cite{wang2022ofa} proposes a unified sequence-to-sequence learning framework to incorporate various VL tasks into the encoder-decoder scheme.
Our work focuses on accelerating the decoder inference for this type of encoder-decoder transformer models.
 
\noindent{\bf Dynamic Early Exit}
Using Dynamic Early Exit (DEE) is a popular strategy to reduce the inference latency of transformer models~\cite{xin2020deebert,liao2021global,Liu_Meng_Zhou_Chen_Xu_2021,liu2020fastbert,zhang-etal-2022-pcee,geng2021romebert,xin2021berxit,zhou2020bert,li2021accelerating}. For example, DeeBERT~\cite{xin2020deebert} and RomeBERT~\cite{geng2021romebert} applies DEE to BERT~\cite{kenton2019bert} based on classification confidence scores from different encoder layers. BERxiT~\cite{xin2021berxit} learns a policy for dynamic early exit. TOKEE~\cite{li2021accelerating} introduces a token-level early exit approach for sequence labelling. However, these encoder-focused approaches cannot be applied to transformer decoders directly, due to the challenges imposed by the auto-regressive mechanism in the decoder models.

DAT~\cite{elbayad2019depth} is one approach tackling decoder early exit. It introduces a halt-and-copy approach, which halts the computation at a layer if the prediction is confident, and copies the feature from shallow decoder layers to deeper layers in later decoding steps when needed. CALM~\cite{schuster2022confident} follows the same halt-and-copy approach for decoder early exit. However, this approach suffers strong semantic misalignment because the semantic information from different decoder layers are not compatible. Thus the later decoding step at deeper layers cannot obtain meaningful features from previous steps, leading to significant accuracy drops. In contrast, our approach dynamically computes the deeper-layer features from earlier steps just-in-time to resolve the semantic misalignment and to achieve higher accuracy than DAT.

\noindent{\bf Multi-exit Models}
The most straightforward way of building multi-exit models is adding deep supervision to each layer \cite{lee2015deeply,teerapittayanon2016branchynet,schwartz2020right}.
Nonetheless, it often degrades the accuracy of the final prediction layer.
To preserve the final layer accuracy, DeeBERT~\cite{xin2020deebert} proposes a two-stage training strategy, in which the final prediction layer and the backbone are trained firstly, and other prediction layers are trained secondly with the rest of the parts frozen.
However, this two-stage training strategy leads to reduced accuracy of shallow layers.
RomeBERT~\cite{geng2021romebert} designs an approach to increase the accuracy of shallow layers using self-distillation and gradient regularization.
BERxiT~\cite{xin2021berxit} uses an alternating training scheme to improve the accuracy of shallow layers. It alternates between two training objectives: the loss of the final layer only and the loss of all layers.
Unlike previous work, we build the multi-exit model by sharing the prediction head among all layers and inserting adaptation modules to align the feature spaces.
Our approach shows the best trade-off between final layer accuracy and shallow layer accuracy.

\noindent{\bf Other Directions for Latency Reduction}
Apart from dynamic early exit, there are attempts in other directions to reduce latency for transformers. For example, knowledge distillation \cite{hinton2015distilling,jiao2020tinybert,lin2022knowledge,sanh2019distilbert} is applied to reduce the model size and latency by distilling information from a large teacher model to a small student model. Model pruning \cite{gordon2020compressing,michel2019sixteen} reduces model size by removing redundant parameters. Non-autoregressive generation \cite{gu2018non,qian2021glancing} avoids the time-consuming step-by-step generation by decoding the predictions in parallel. These directions are orthogonal to dynamic early exit, hence they are not our focus.

\section{Approach}
\label{sec:method}

We propose \oursnospace, a dynamic early exit on decoder approach to accelerate encoder-decoder transformer models for VL tasks.
Specifically, we leverage confidence-based step-level dynamic early exit to decide which decoder layer to exit based on how confident we are at each decoding step.
At training, we train our multi-exit model with deep supervision~\cite{lee2015deeply}, where the output features of each decoder layer are input to a shared generation head and supervised using the ground truth.
At inference, we apply dynamic early exit on the auto-regressive decoder. At each decoding step, the model decides how many decoder layers to use based on its confidence about the output token – hence different number of layers may be used at different decoding steps.
In the follow sections, we first introduce the auto-regressive decoding process and the challenge of semantic misalignment in dynamic early exit on decoder in~Section \ref{sec:background}.
Then we describe our multi-exit model architecture and our training strategy in~Section \ref{sec:model}.
Finally we show how we resolve the semantic misalignment problem with just-in-time computation of decoder features in~Section \ref{sec:step-level}.

\subsection{Background}
\label{sec:background}
\paragraph{Auto-Regressive Decoding}
At inference, decoder typically generates the prediction in an auto-regressive decoding way,
\ie, decoder generates tokens step-by-step and the generated token in each step is conditioned on the previously generated tokens.
Theoretically, all the previous tokens are supposed to be input to the decoder to generate the current token, 
which would cause redundant computation for the previous tokens as their features have been computed at previous decoding steps.
In common practice, to reduce redundant computation, the key-value features in the multi-head self-attention layers are all saved and provided for later steps.
This practice decreases computation complexity and reduces inference latency, by avoid re-computing key-value features of earlier decoder steps at later steps.

\noindent{\bf Semantic Misalignment}
In step-level dynamic early exit, 
each decoding step can use a different number of decoder layers.
As a result, the past key-value features may not always be available for every layer.
This misalignment makes it difficult to implement the step-level dynamic early exit, 
as it cannot retrieve the cached key-value features from previous steps when the current step uses deeper layers.
One option is to copy shallower-layer key-value features to deeper-layers \cite{elbayad2019depth}.
However, the deeper-layer features encode higher-level semantics compared to shallower-layer features. A mixture of them across decoding steps will cause semantic misalignment and undermine the generative ability of the model.
An easy workaround is to constrain the model to always exit at the same decoder layer,
but this would upset our observation that some tokens are harder to generate than others.
In experiments, we will show that this constrained approach is not desirable in terms of accuracy and latency.
While we have to stick to step-level dynamic early exit and solve semantic misalignment, 
pre-computing the deeper-layer key-value features is not efficient because we do not know  how many layers the following steps will use.
To address this issue, we do step-level dynamic early exit with just-in-time computation, see Section \ref{sec:step-level}.

\begin{figure}[t]
    \centering
    \includegraphics[width=0.9\linewidth]{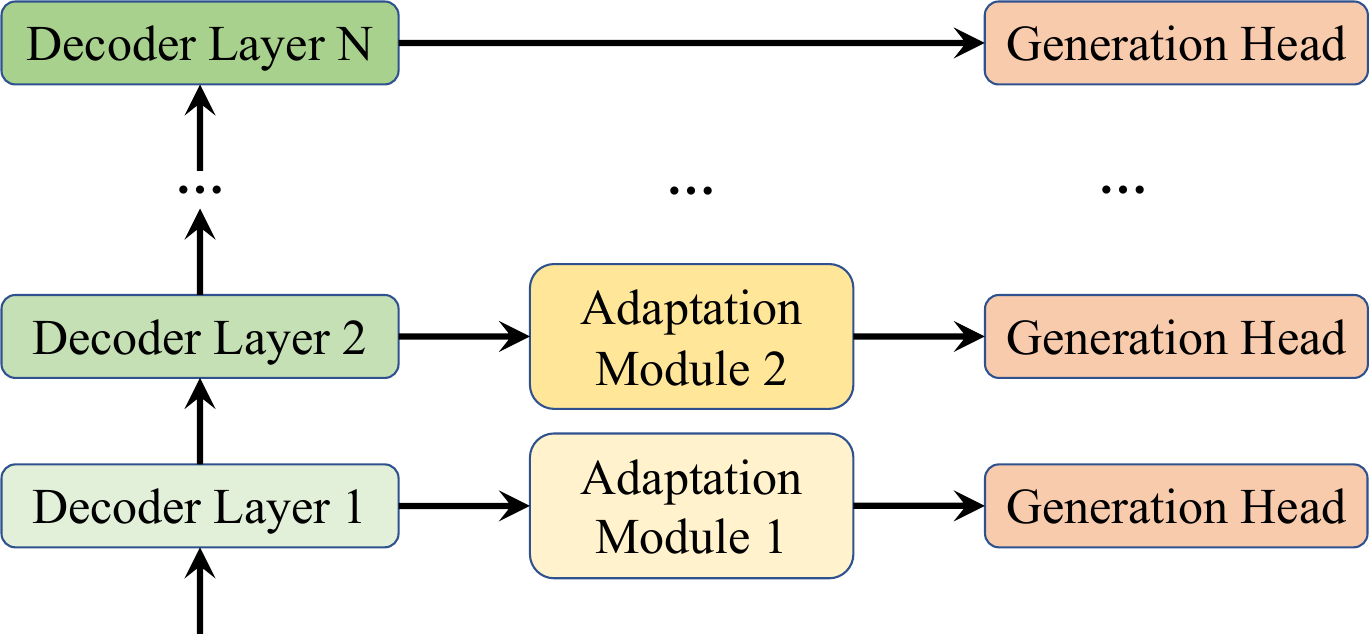}
     \caption{Decoder architecture of our multi-exit model. We share the generation head across different decoder layers and insert adaptation modules between early decoder layers and the generation head.}
    \label{fig:model_arch}
\end{figure}

\subsection{Multi-exit Model}
\label{sec:model}

To perform step-level dynamic early exit, it is crucial to have a multi-exit model to ensure each decoder layer is capable of generating plausible predictions.
So we introduce our multi-exit model here before moving on to how we do step-level dynamic early exit.

\noindent{\bf Model Architecture}
In our multi-exit encoder-decoder transformer model, we have a generation head that maps decoder features into tokens.
In contrast to existing work \cite{geng2021romebert,xin2020deebert,xin2021berxit}, we share the generation head across different decoder layers to share the common generation knowledge among different decoder layers, which strengthens the generative ability of shallow decoder layers.
In addition, we insert separate adaptation modules between the shallow decoder layers and the generation head to adapt the features from shallow decoder layers to the semantic space of features from the final decoder layer (see Figure \ref{fig:model_arch}).
Specifically, the adaptation module is composed of a linear layer followed by layer normalization.

\noindent{\bf Model Training}
To train the multi-exit model, the most straightforward way is to add deep supervision \cite{lee2015deeply} after outputs of each decoder layer as follows 
\begin{equation}
\label{eq:loss_ave}
    \mathcal{L}_{\mathit{avg}}= \frac{1}{N}\sum_{n=1}^{N} \mathcal{L}_n,
\end{equation}
where $N, \mathcal{L}_n, \mathcal{L}_{\mathit{avg}}$ correspond to the total number of decoder layers, the loss for the $n$-th decoder layer, and the average loss across all decoder layers, respectively.
However, this approach does not optimize the model for the final decoder layer solely.
As a result, the model suffers from degraded accuracy of the final decoder layer, which will cap the accuracy of our approach.
To address this issue, we emphasize the loss of the final layer so as to maintain high accuracy for the final decoder layer.
To this end, we add the final decoder layer loss to the training objective as follows
\begin{equation}
\label{eq:loss_total}
    \mathcal{L} = \mathcal{L}_{\mathit{avg}} + \mathcal{L}_N.
\end{equation}

\begin{algorithm}[t]
\caption{Step-level dynamic early exit with just-in-time computation}
\label{alg:step-level}
\begin{algorithmic}[1]
    \REQUIRE Current decoding step $i$, saved past key-value features $\mathcal{P} = \{\mathbf{p}_{n}^{i'}\}$, saved hidden states $\mathcal{H} = \{\mathbf{h}_{n}^{i'}\}$, decoder layers $\mathcal{D} = \{\mathop{D}_{n}\}$, the number of decoder layers $N$, confidence score threshold \thnospace.
    \ENSURE Decoded token output $t^{i}$.
    \FOR{$n=1$ \TO $N$}
      \STATE Get saved past key-value features $\mathbf{p}_{n}^{1:j}$ and hidden states $\mathbf{h}_{n-1}^{j+1:i}$.
      \STATE Feed $\mathbf{p}_{n}^{1:j}$ and $\mathbf{h}_{n-1}^{j+1:i}$ into $\mathop{D}_{n}$ to compute $\mathbf{p}_{n}^{j+1:i}, \mathbf{h}_{n}^{j+1:i}, t_{n}^{i}$ with confidence score $c_{n}^{i}$.
      \STATE Save $\mathbf{p}_{n}^{j+1:i}$ to $\mathcal{P}$ and $\mathbf{h}_{n}^{j+1:i}$ to $\mathcal{H}$.
      \IF{$c_{n}^{i} > \tau$}
          \STATE Set $t^{i}$ to $t_{n}^{i}$ and terminate the \textbf{for} loop.
      \ENDIF
    \ENDFOR
\end{algorithmic}
\end{algorithm}

\subsection{Step-Level Dynamic Early Exit with Just-in-Time Computation}
\label{sec:step-level}

We perform step-level dynamic early exit at inference on top of the multi-exit model.
To avoid semantic misalignment and improve efficiency, we design an algorithm to compute the past key-value features just-in-time.
For step $i$ and decoder layer $n$, we denote the decoder layer $n$ as $\mathop{D}_{n}$, the past key-value features as $\mathbf{p}_n^i$, the decoded token output as $t_{n}^{i}$, the corresponding confidence score as $c_{n}^{i}$, and the confidence score threshold as \thnospace.
We use colon separated numbers to denote intervals, \eg, $i:j$ denotes the decoding steps from $i$ to $j$ (inclusive).
Apart from $\mathbf{p}_n^i$, we also save any output hidden states $\mathbf{h}_n^i$ of $\mathop{D}_{n}$ at step $i$ if it is computed. 

\begin{figure}[t]
    \centering
    \includegraphics[width=0.63\linewidth]{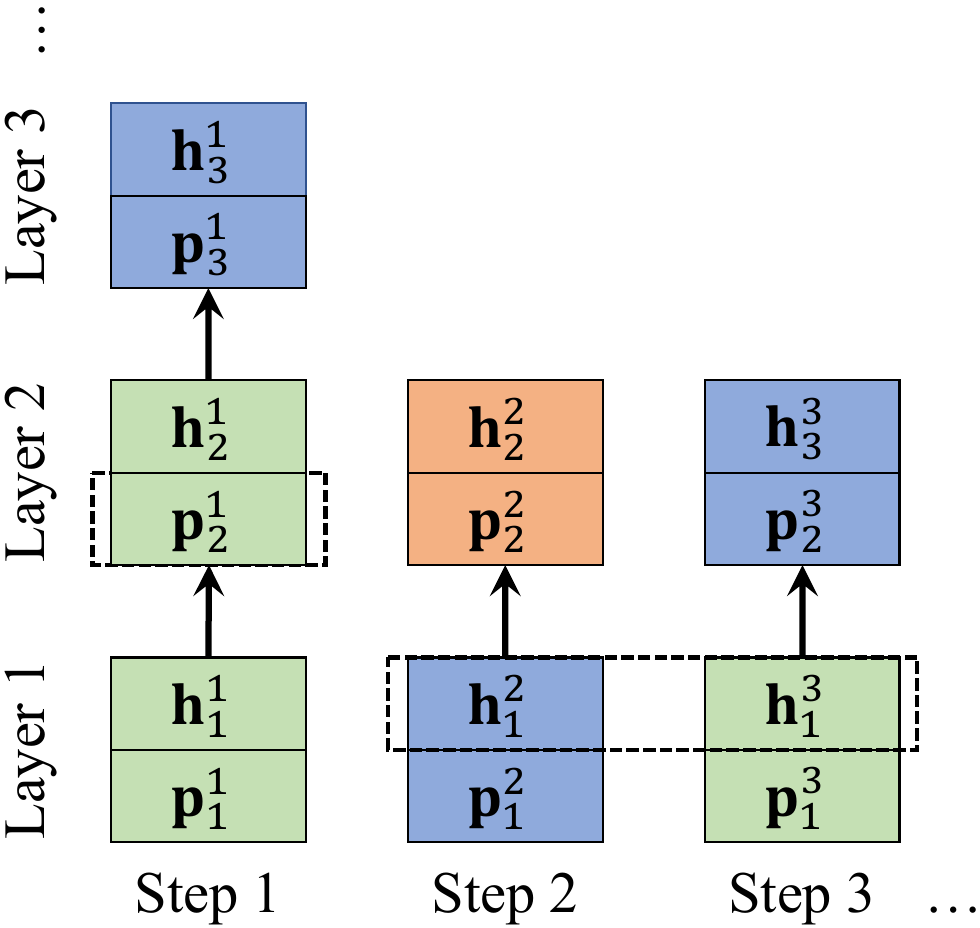}
    \caption{Step-level dynamic early exit with just-in-time computation.
    Light blue boxes: the decoder layers where the model exits.
    Light green boxes: the internal decoder layers. 
    Light orange boxes: the layer used for just-in-time computation.
    Features/hidden states in the dashed boxes are inputs to decoder layer 2 at decoding step 3.}
    \label{fig:step-level}
\end{figure}

\begin{table*}[t]
    \centering
    \scalebox{0.84}{
    \begin{tabular}{l | l l l | l l l}
        \toprule
        \multirow{2}{*}{} & \multicolumn{3}{l|}{DocVQA} & \multicolumn{3}{l}{OCR-VQA} \\
        \cline{2-7}
        & ANLS$\uparrow$ & Dec. Latency$\downarrow$ & Tot. Latency$\downarrow$ & Accuracy$\uparrow$ & Dec. Latency$\downarrow$ & Tot. Latency$\downarrow$ \\
        \midrule
        Original-b & 81.5 & 104.3 & 124.6 & 68.4 & 109.7 & 125.6 \\
        \midrule
        DAT-b \cite{elbayad2019depth} & 71.4 & \ \ 90.9 & 111.6 & 60.3 & 105.3 & 121.4 \\
        SLEX-b & 81.4 & \ \ 90.1 & 111.4 & \bf 68.3 & 109.0 & 124.7 \\
        FTEX-b & 81.2 & \ \ 47.1 & \ \ 67.4 & 67.1 & \ \ 53.8 & \ \ 70.3 \\
        \oursnospace-b & {\bf 81.9}\textsubscript{+0.4} & \ \ {\bf 46.1}\textsubscript{-55.8\%} & \ \ {\bf 66.5}\textsubscript{-48.6\%} & 68.1\textsubscript{-0.3} & \ \ {\bf 52.4}\textsubscript{-52.2\%} & \ \ {\bf 68.5}\textsubscript{-45.5\%} \\
        \midrule
        \midrule
        Original-L & 83.5 & 181.5 & 216.3 & 70.1 & 202.5 & 229.6 \\
        \midrule
        DAT-L \cite{elbayad2019depth} & 74.3 & 134.0 & 169.9 & 63.0 & 166.0 & 194.2 \\
        SLEX-L & 83.7 & 154.3 & 190.6 & 69.6 & 111.5 & 139.8 \\
        FTEX-L & 83.1 & \ \ 58.6 & \ \ 91.5 & 68.6 & \ \ 79.6 & 108.1 \\
        \oursnospace-L & {\bf 83.8}\textsubscript{+0.3} & \ \ {\bf 49.2}\textsubscript{-72.9\%} & \ \ {\bf 82.8}\textsubscript{-61.7\%} & {\bf 69.7}\textsubscript{-0.4} & \ \ {\bf 79.2}\textsubscript{-60.9\%} & {\bf 107.5}\textsubscript{-53.2\%} \\
        \bottomrule
    \end{tabular}
    }
    \caption{Accuracy and latency (in ms) on DocVQA and OCR-VQA validation sets. The best results are in {\bf bold face}. The percentage reductions are \wrt the original model.}
    \label{tab:main_results_docvqa_and_ocrvqa}
\end{table*}

As shown in Algorithm \ref{alg:step-level}, for each decoding step, we go through the decoder one layer per iteration.
At decoding step $i$, first we prepare saved past key-value features $\mathbf{p}_{n}^{1:j}$ and hidden states $\mathbf{h}_{n-1}^{j+1:i}$ (for the decoding steps where the key-value features are absent), where $j$ ($< i$) corresponds to the sequence length of saved past key-value features for $\mathop{D}_{n}$, see line 2 in Algorithm \ref{alg:step-level}.
Next we feed $\mathbf{p}_{n}^{1:j}$ and $\mathbf{h}_{n-1}^{j+1:i}$ into $\mathop{D}_{n}$ to compute key-value features $\mathbf{p}_{n}^{j+1:i}$, hidden states $\mathbf{h}_{n}^{j+1:i}$, decoded token output $t_{n}^{i}$, and the corresponding confidence score $c_{n}^{i}$, see line 3 in Algorithm \ref{alg:step-level}.
We save these newly computed $\mathbf{p}_{n}^{j+1:i}$ and $\mathbf{h}_{n}^{j+1:i}$ for future use, see line 4 in Algorithm \ref{alg:step-level}.
Taking the decoding process in Figure \ref{fig:step-level} as an example, at decoding step 3 when the model is about to enter layer 2, the past key-value features $\mathbf{p}_{2}^{1}$ are available but $p_{2}^{2}$ are absent, so $\mathbf{p}_{2}^{1}$ along with the saved hidden states $\mathbf{h}_{1}^{2:3}$ will be fed into decoder layer 2.
We repeat the same process for every decoder layer until the predicted confidence score $c_{n}^{i}$ is larger than a threshold \thnospace, where $c_{n}^{i}$ is computed by the classification score after softmax, see line 5-6 in Algorithm \ref{alg:step-level}.
Note that although the deeper-layer features are computed for the previous decoding steps, the previous token outputs will not be updated with those features, because each token is supposed to be dependent on the past and any change in the previous tokens will break the dependency.

One may notice that our approach assumes the $\mathbf{h}_{n-1}^{j+1:i}$ available at decoding step $i$.
This is assured by our per-layer traversal - the hidden states are always computed and saved at the previous decoder layer.

\section{Experiments}
\label{sec:exp}

\begin{table*}[t]
    \centering
    \scalebox{0.84}{
    \begin{tabular}{l | l l l | l l l}
        \toprule
        \multirow{2}{*}{} & \multicolumn{3}{l|}{ST-VQA} & \multicolumn{3}{l}{Text-VQA} \\
        \cline{2-7}
        & ANLS$\uparrow$ & Dec. Latency$\downarrow$ & Tot. Latency$\downarrow$ & Accuracy$\uparrow$ & Dec. Latency$\downarrow$ & Tot. Latency$\downarrow$ \\
        \midrule
        Original-b & 69.7 & \ \ 71.9 & \ \ 88.6 & 61.1 & \ \ 71.7 & \ \ 89.0 \\
        \midrule
        DAT-b \cite{elbayad2019depth} & 62.8 & \ \ 57.3 & \ \ 74.4 & 53.3 & \ \ 64.4 & \ \ 83.3 \\
        SLEX-b & 69.8 & \ \ 60.4 & \ \ 77.3 & 59.6 & \ \ 51.9 & \ \ 68.9 \\
        FTEX-b & 69.5 & \ \ 41.7 & \ \ 59.0 & 60.0 & \ \ 45.5 & \ \ 62.4 \\
        \oursnospace-b & {\bf 69.9}\textsubscript{+0.2} & \ \ {\bf 33.5}\textsubscript{-53.4\%} & \ \ {\bf 50.1}\textsubscript{-43.5\%} & {\bf 61.0}\textsubscript{-0.1} & \ \ {\bf 43.5}\textsubscript{-39.3\%} & \ \ {\bf 61.4}\textsubscript{-31.0\%} \\
        \midrule
        \midrule
        Original-L & 70.3 & 136.5 & 164.2 & 63.1 & 136.5 & 165.5  \\
        \midrule
        DAT-L \cite{elbayad2019depth} & 62.8 & \ \ 85.5 & 112.4 & 54.6 & 111.1 & 140.5 \\
        SLEX-L & 70.2 & \ \ 85.0 & 114.5 & 61.3 & \ \ 93.5 & 122.9 \\
        FTEX-L & 70.4 & \ \ 65.9 & \ \ 96.4 & 61.8 & \ \ 79.3 & 108.9 \\
        \oursnospace-L & {\bf 71.5}\textsubscript{+1.2} & \ \ {\bf 50.0}\textsubscript{-63.4\%} & \ \ {\bf 78.5}\textsubscript{-52.2\%} & {\bf 63.6}\textsubscript{+0.5} & \ \ {\bf 72.1}\textsubscript{-47.2\%} & {\bf 102.7}\textsubscript{-37.9\%} \\
        \bottomrule
    \end{tabular}
    }
    \caption{Accuracy and latency (in ms) on ST-VQA and Text-VQA validation sets. The best results are in {\bf bold face}. The percentage reductions are \wrt the original model.}
    \label{tab:main_results_stvqa_and_textvqa}
\end{table*}

We evaluate \ours on two state-of-the-art encoder-decoder models: LaTr++~\cite{biten2022latr} and OFA~\cite{wang2022ofa} with various vision-language tasks. We do auto-regressive prediction for all the tasks.

\subsection{\ours on LaTr++}
\label{sec:exp:subsec:latr}
LaTr~\cite{biten2022latr} is the state-of-the-art approach for text-based visual question answering (text-VQA). 
LaTr uses multi-modal encoder-decoder transformer models with OCR text, layout, and visual features as inputs.
We improve LaTr by using a better vision backbone and adding better unsupervised pre-training tasks, see Section \ref{sec_sup:pretrain} for more details.
We refer to the improved LaTr as LaTr++ here.
Following LaTr, we focus on the text-VQA task.

\begin{table*}[t]
    \centering
    \small
    \scalebox{0.86}{
    \begin{tabular}{l | c c c c c c c c c}
        \toprule
        \multirow{2}{*}{} & \multicolumn{4}{c|}{VQA} & \multicolumn{5}{c}{RefCOCO} \\
        \cline{2-10}
        & test-dev$\uparrow$ & test-std$\uparrow$ & Dec. Lat.$\downarrow$ & \multicolumn{1}{c|}{Tot. Lat.$\downarrow$} & val $\uparrow$ & testA $\uparrow$ & testB $\uparrow$ & Dec. Lat.$\downarrow$ & Tot. Lat.$\downarrow$ \\
        \midrule
        OFA & 79.3 & 79.4 & 753.5 & \multicolumn{1}{c|}{811.3} & 90.6 & 92.5 & 85.9 & 132.8 & 187.1\\
        \ours & 79.0 & 79.1 & 480.7 & \multicolumn{1}{c|}{538.5} & 90.2 & 92.4 & 85.1 & 79.5 & 133.8\\
        \midrule
        \multirow{2}{*}{} & \multicolumn{5}{c|}{RefCOCO+} & \multicolumn{4}{c}{RefCOCOg} \\
        \cline{2-10}
        & val $\uparrow$ & testA $\uparrow$ & testB $\uparrow$ & Dec. Lat.$\downarrow$ & \multicolumn{1}{c|}{Tot. Lat.$\downarrow$} & val-u $\uparrow$ & test-u $\uparrow$ & Dec. Lat.$\downarrow$ & Tot. Lat.$\downarrow$ \\
        \midrule
        OFA & 85.7 & 89.9 & 78.6 & 142.7 & \multicolumn{1}{c|}{197.9} & 87.2 & 87.6 & 132.5 & 187.6 \\
        \ours & 85.3 & 89.6 & 77.9 & 61.6 & \multicolumn{1}{c|}{117.9} & 87.0 & 87.4 & 83.0 & 138.1 \\
        \bottomrule
    \end{tabular}
    }
    \caption{Accuracy and latency (in ms) of OFA and \ours on various multi-modal tasks. The large size model is used here.}
    \label{tab:main_results_ofa}
\end{table*}

\subsubsection{Settings}

We evaluate on four text-VQA datasets: DocVQA \cite{mathew2021docvqa}, OCR-VQA \cite{mishra2019ocr}, ST-VQA \cite{biten2019scene}, and TextVQA \cite{singh2019towards}, using accuracy and latency as the metric.
For accuracy, we follow the standard protocol to report the metrics on each dataset, \ie, Average Normalized Levenshtein Similarity (ANLS) \cite{biten2019scene,mathew2021docvqa} for DocVQA and ST-VQA, and accuracy of exact text match between groundtruth and prediction for OCR-VQA and TextVQA. For latency, we report both the total inference latency and the decoder-only latency, as our approach only affects the decoder inference. The latency is measured \wrt wall-clock time on the same machine which has 1 Nvidia A100 GPU with 40GB memory. All approaches are implemented in Pytorch~\cite{paszke2017automatic} with Huggingface~\cite{wolf2019huggingface}. To measure the most accurate per sample latency, we use batch size 1 in inference to avoid unnecessary padding.
See Section \ref{sec_supp:settings} for more details.

\noindent{\bf Baselines}
We compare \ours to the original model, the SOTA approach DAT~\cite{elbayad2019depth}, and two strong baselines SLEX and FTEX we proposed and built:
\begin{itemize}
    \item Original: the vanilla LaTr++ model, on which no early-exit or deep-supervision is applied.
    
    \item DAT~\cite{elbayad2019depth}: DAT is the state-of-the-art decoder speed-up algorithm. At the step when the model exits at a deeper layer, it simply copies the features of the shallow layer from previous steps to all deeper layers. 
    
    \item Sequence-level early exit (SLEX): the decoder always exits at layer $m$ at each decoding step. $m$ is chosen by the accumulated confidence score of the entire sequence. More precisely, for each decoder layer, SLEX needs to infer all decoding steps to get the accumulated confidence score, which makes this baseline unpractical.
    
    \item First-token early exit (FTEX): the decoder always exit at layer $m$ at each decoding step. Unlike SLEX, $m$ is chosen based on the confidence score of the first token, which makes FTEX more practical than SLEX because FTEX only needs to infer the first decoding step to make the decision.
\end{itemize}

In our experiments, for fair comparisons, SLEX, FTEX, and DAT use the same multi-exit model trained for \ours, which improves the accuracy of the shallow layers.

\noindent{\bf Implementation Details}
We evaluate our approach \ours on both base (-b) and large (-L) variations of LaTr++. The base version has 12 encoder and 12 decoder layers, and the large version has 24 encoder and 24 decoder layers. We follow LaTr \cite{biten2022latr} to do pre-training first and fine-tuning later. We add deep supervision loss in Eq.~(\ref{eq:loss_total}) in both pre-training and fine-tuning. The confidence score threshold \th is selected using cross-validation, specifically, 0.99 on DocVQA and 0.95 for other three datasets. See Section \ref{sec_supp:implement_detail} for more details.

\subsubsection{Results}
\label{sec:exp:subsec:latr:subsubsec:results}
Table~\ref{tab:main_results_docvqa_and_ocrvqa} and Table~\ref{tab:main_results_stvqa_and_textvqa} show the comparisons of accuracy and latency among \ours and baseline approaches.

\begin{figure*}[t]
     \centering
     \begin{subfigure}{0.33\linewidth}
         \centering
         \includegraphics[width=\linewidth]{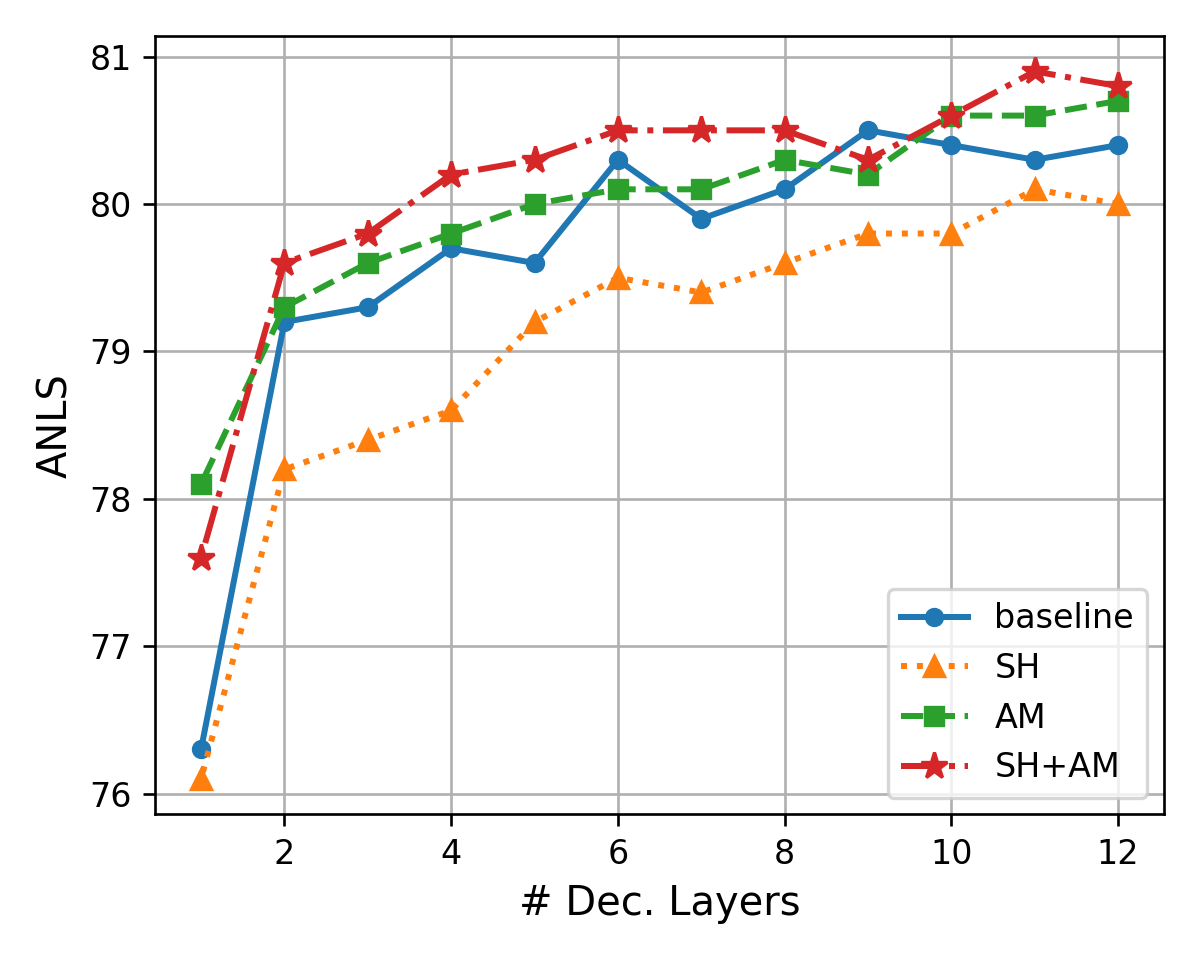}
         \caption{Ablation on model architecture.}
         \label{fig:abl_arch}
     \end{subfigure}
     \hfill
     \begin{subfigure}{0.33\linewidth}
         \centering
         \includegraphics[width=\linewidth]{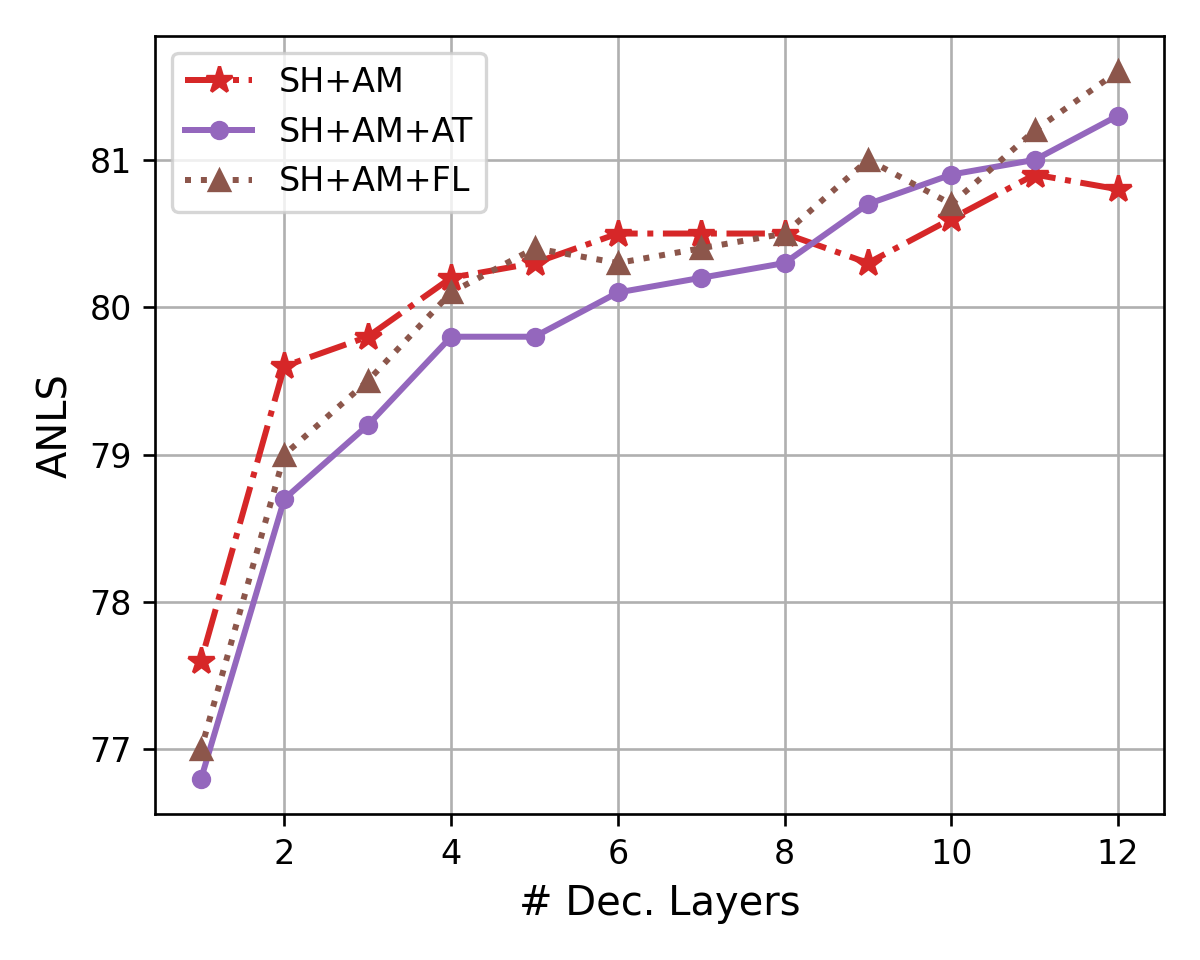}
         \caption{Ablation on training objective.}
         \label{fig:abl_train}
     \end{subfigure}
     \hfill
     \begin{subfigure}{0.33\linewidth}
         \centering
         \includegraphics[width=\linewidth]{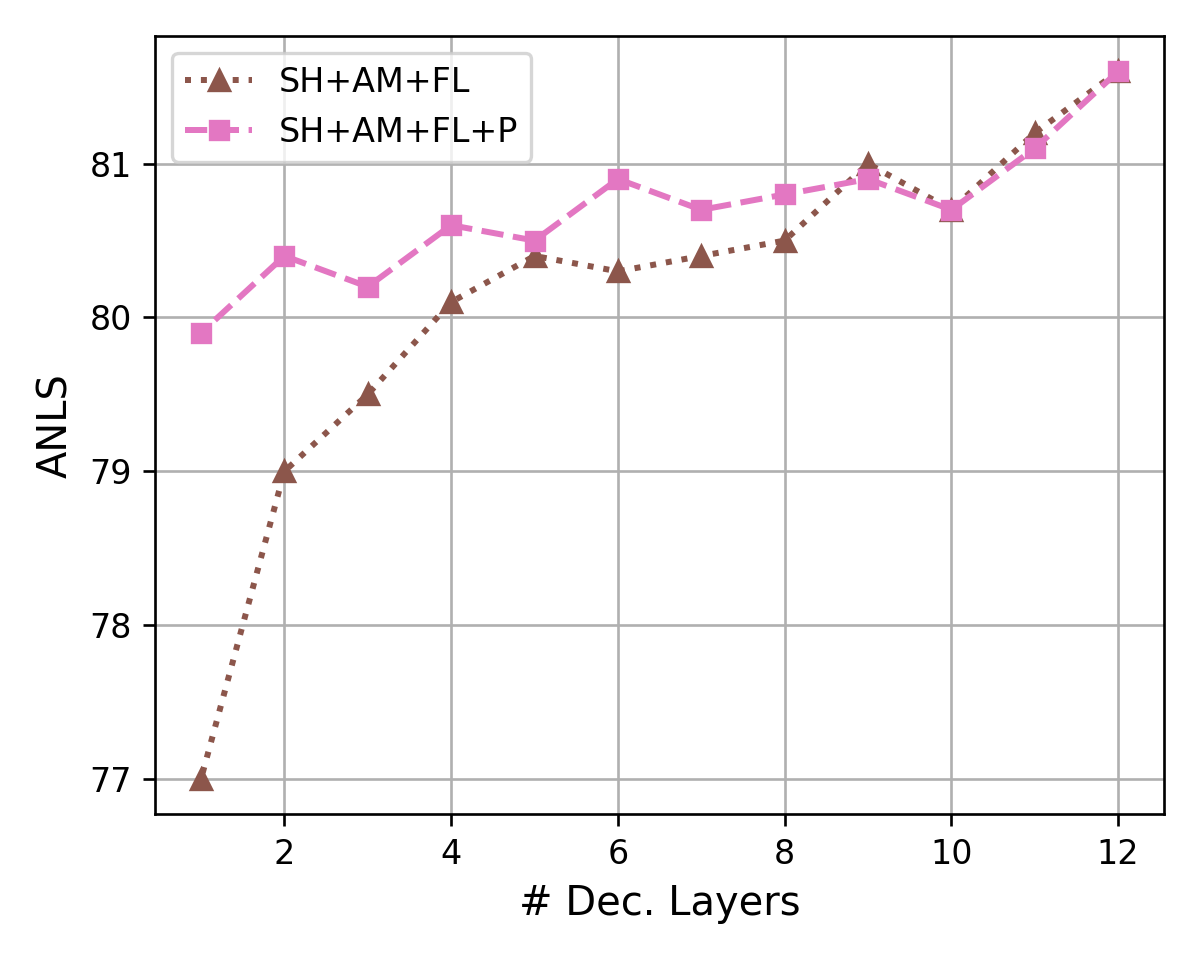}
         \caption{Ablation on pre-training.}
         \label{fig:abl_pre-train}
     \end{subfigure}
    \caption{Ablation studies on the techniques for building and training the multi-exit model. The figure plots the ANLS when exiting at the specific decoder layer. SH: share generation head, AM: adaptation module, AT: alternating training, FL: Additional final layer loss, P: Pre-training with deep supervision, Baseline: the baseline model with unshared generation heads and without the adaptation module.}
    \label{fig:multi-exit}
\end{figure*}

Our approach shows excellent performance compared to the original model. It consistently reduces the inference latency for both base and large variations, 
while maintaining the evaluation accuracy on all benchmark datasets. The latency reduction on decoder is between 40\% and 73\% across all model and dataset combinations. Specifically, on DocVQA, \ours reduces the decoder latency on the larger variation from 181.5ms to 49.2ms, achieving a large 72.9\% reduction, while its ANLS is 0.3 higher than the original LaTr++. 
Comparing to other baseline approaches, \ours always outperforms them with clear margins. DAT~\cite{elbayad2019depth} reduces the decoder latency slightly, but it suffers from major accuracy degradation, due to the semantic misalignment introduced by the copy mechanism. SLEX can maintain high accuracy as it makes the decision based on the entire sequence, but its latency improvement is minor compared to \ours. FTEX can reach significant inference acceleration as it can decide to use a shallow layer after the first decoding step. However, it often sacrifices more accuracy because the layer with the maximum first token confidence might not have the best generation for the entire sequence. In contrast, our approach makes exit decisions at each layer and each step, and recomputes the deeper features when necessary, which helps it achieve the best accuracy and latency comparing to all other approaches.

Notice that there is usually more the latency reduction on the large model, because the large model has more decoder layers and early exit still happens at very shallow layers instead of going deeper.
In addition, our approach can improve the accuracy of the vanilla LaTr++ in most cases,
because our multi-exit model with deep supervision pre-training significantly improves the accuracy for shallow layers (see Section \ref{sec:exp:subsec:ablation}), and \ours often chooses the layer with the best generative ability based on the confidence scores.
In fact, shallow layers can have better predictions than deeper layers on certain examples.
If we can choose which layer to make the prediction according to groundtruth, the base model can obtain ANLS 85.0 on DocVQA.

\subsection{\ours on OFA}
\label{sec:exp:subsec:ofa}
OFA~\cite{wang2022ofa} is a sequence-to-sequence framework that unifies multiple modalities. It can address multiple VL tasks with a single paradigm. The encoder and decoder are chosen based on the task. 

\noindent{\bf Experimental Setup}
Following \cite{wang2022ofa}, we evaluate \ours with OFA on various multi-modal downstream tasks,
specifically, VQAv2~\cite{balanced_vqa_v2} for VQA, and RefCOCO/RefCOCO+/RefCOCOg \cite{yu2016modeling,mao2016generation} for referring expression comprehension.
We also report the accuracy, the total inference latency, and the decoder-only latency and compare \ours to the baseline model, as described in Section \ref{sec:exp:subsec:latr}. For each dataset, the latency is averaged on samples from all splits. Here we only compare to the original OFA model. We use the original pre-trained OFA model and the same fine-tuning procedure as in \cite{wang2022ofa} to reproduce the OFA results and train \oursnospace.
We do not pre-train the model with deep supervision due to its overwhelming computational costs. We use the large size OFA model and the threshold \th is chosen via cross-validation, \ie, 0.96 for VQA and 0.1 for RefCOCO/RefCOCO+/RefCOCOg.

\noindent{\bf Results}
The accuracy and latency of the original OFA and \ours are shown in Table \ref{tab:main_results_ofa}. The results of OFA are reproduced using the official code, which are very close to the reported numbers. 
Again \ours consistently reduces the decoder inference latency with marginal accuracy drops. Specifically, it achieves an average 36.2\% and 44\% decoder latency reduction on the VQA task and the referring expression comprehension task respectively.
In addition, even without deep-supervision pre-training, \ours obtains comparable accuracy compared to the original OFA.
The accuracy of \ours should be boosted if we do deep-supervision pre-training for OFA as well.
These results demonstrate that our approach can be generalized to different encoder-decoder transformer models and various VL tasks.

\begin{figure}[ht]
    \centering
    \includegraphics[width=0.74\linewidth]{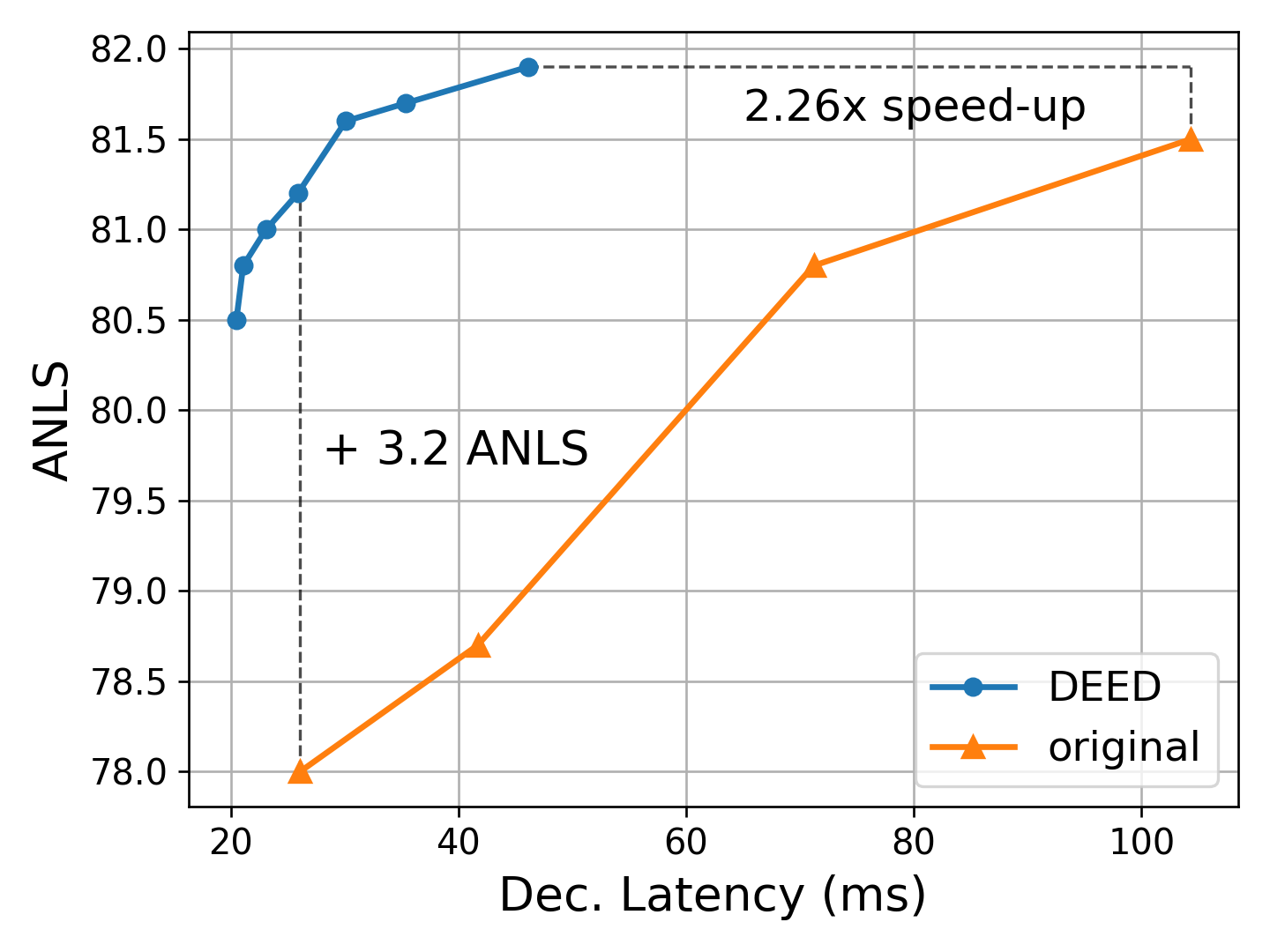}
    \caption{ANLS against decoder latency for \ours and the original model on the DocVQA validation set. \ours are obtained by tuning the confidence threshold \th. The original model is trained with different numbers of decoder layers.}
    \label{fig:trade-off}
\end{figure}

\subsection{Ablation Study}
\label{sec:exp:subsec:ablation}
We study the contribution of each component in \oursnospace. All ablation studies are conducted on DocVQA with LaTr++ base variation.

\noindent{\bf Model Architecture}
We inspect the effect of shared generation head (SH) and the adaptation module (AM) for multi-exit model. We compare the models of using SH only, using AM only, and using both (SH + AM), to the baseline model trained with unshared generation heads without the adaptation module. 
We use $\mathcal{L}_{\mathit{avg}}$ in Eq. \ref{eq:loss_ave} for training and we do not do deep-supervision pre-training.
Figure~\ref{fig:abl_arch} shows results of different approaches.
By using both the shared generation head and the adaptation module, the model achieves consistently better accuracy than the baseline, except for the 9-th layer. Notice that the improvement on the first layer is the greatest ($>$1\%), which hugely contributes to the overall latency reduction as more examples can exit at layer 1 without sacrificing the accuracy. However, without the adaptation module, the shared generation head has inferior performance due to the mis-alignment between the generation head and the intermediate features for generation.

\noindent{\bf Training Objective}
In Figure \ref{fig:abl_train}, we visualize the ANLS of models trained with the vallina deep supervision $\mathcal{L}_{\mathit{avg}}$ in Eq. \ref{eq:loss_ave}, alternating training (AT) \cite{xin2021berxit}, and the additional final layer loss (FL) $\mathcal{L}_N$ in Eq. \ref{eq:loss_total}. We can see both AT and FL can improve the accuracy of the deep ($>$ 8) layers, which helps \ours achieve the same or even better accuracy compared to the original model.
FL gives better accuracy for most layers than AT, which confirms the effectiveness of our proposed training objective.

\begin{figure}[t]
    \centering
    \includegraphics[width=\linewidth]{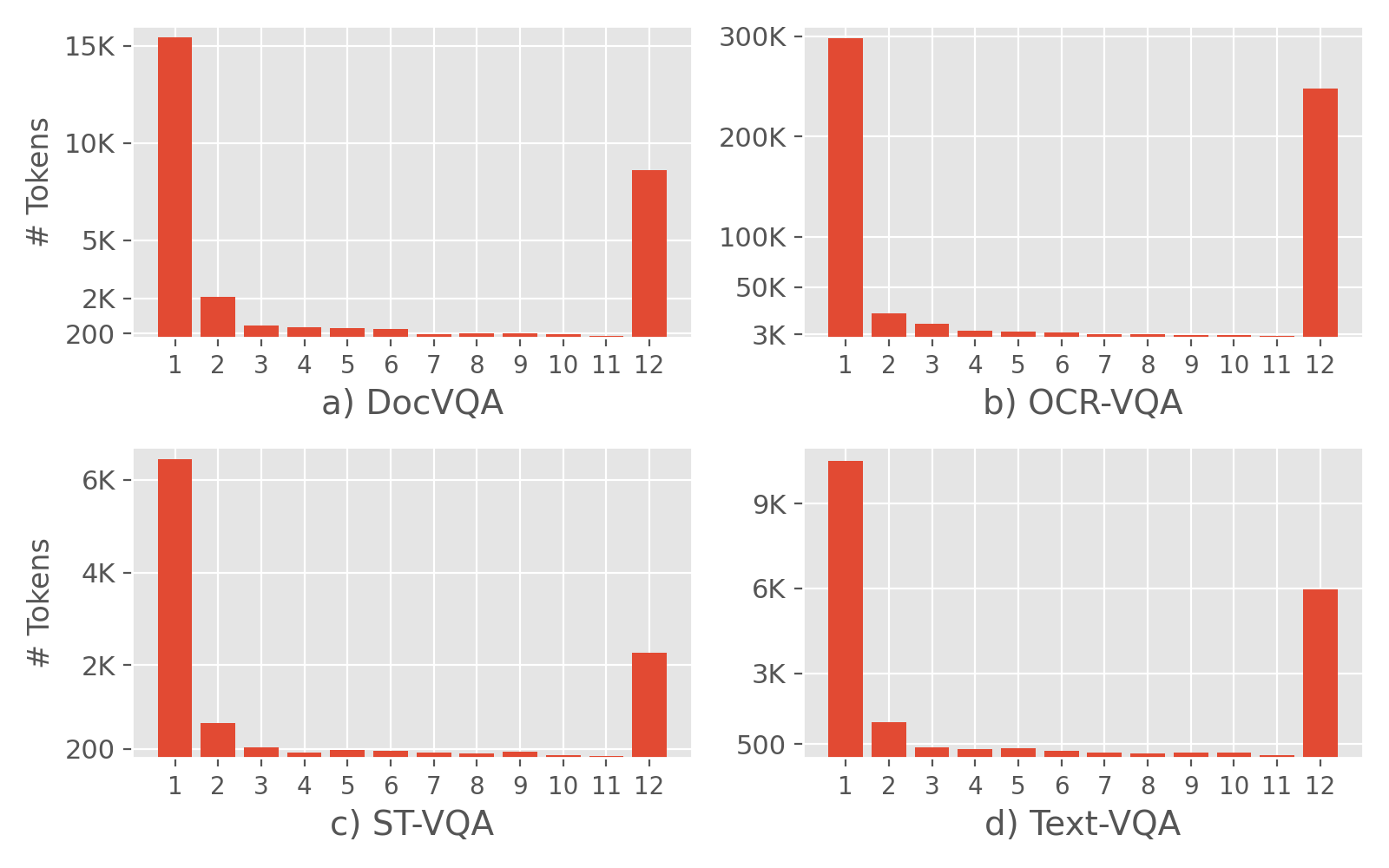}
    \caption{Histogram of layer at which our model exits when evaluated on the validation set of different VQA datasets.}
    \label{fig:layerusage}
\end{figure}

\noindent{\bf Pre-training}
In our experiments, we found that pre-training the model with deep supervision can significantly improve the accuracy of the shallow layers, as shown in Figure \ref{fig:abl_pre-train}. The {\color{magenta}magenta curve} is the model pre-trained with the deep supervision while the {\color{RawSienna}brown curve} is the one without. Pre-training with deep supervision increases the accuracy of the first layer by 3\%. It also consistently increases the ANLS between layer 2 and layer 8. We argue that deep supervision during the pre-training stage helps the model learn strong generative ability in the shallow layers.

\noindent{\bf Threshold \th}
\ours can realize different trade-offs between the accuracy and latency by tuning the confidence score threshold $\tau$, to fit in different use cases without retraining the model. 
In Figure \ref{fig:trade-off}, we visualize the decoder latency and ANLS of \ours \wrt different thresholds ([0.5, 0.99]). We compare \ours to the original LaTr++ trained with 2, 4, 8, and 12 decoder layers. When using the threshold 0.99, our approach reaches the highest ANLS score of 81.9, which exceeds the vanilla 12-layer LaTr++, while achieving 2.26X decoder speed-up. At the other end of the spectrum, \ours can reduce the decoder latency to 25.9ms (4.03X speed-up \vs 12-layer LaTr++) with ANLS score of 81.2. In contrast, to reduce the latency to 26ms, the original model can only use 2 decoder layers, resulting in a significant 3.2 ANLS drop compared to \oursnospace.

\noindent{\bf The Distribution of Tokens Exiting at Each Layer}
We visualize the distribution of tokens exit at each layer on four text-VQA datasets in Figure \ref{fig:layerusage}. The majority of the tokens exit at the first layer, then the last layer. Only a small amount of tokens exit at middle layers.
This shows that the model exits at shallow layers for easy predictions, which aligns with our observation that most samples do not need all decoder layers during inference.
In addition, from Figure 4 in the main paper, the second most samples are hard samples that can be correctly predicted by the final decoder layer or even the final decoder layer fails,
so the second peak of the histogram appears at the final decoder layer (\ie, decoder layer 12).

\section{Conclusions}
We propose \oursnospace, a multi-exit model with step-level dynamic early exit on decoder for encoder-decoder transformer model acceleration. \ours leverages confidence-based step-level dynamic early exit to reduce the computation at each decoding step. To improve the accuracy when exiting at shallow layers, we build a multi-exit model leveraging multiple techniques including deep supervision, shared generation head, adaptation modules, and emphasizing the learning of the final decoder layer. We apply our approach to two state-of-the-art encoder-decoder transformer models. 
Results on various vision-language tasks and datasets show that our approach significantly reduces the inference latency with comparable or even higher accuracy compared to baselines.

\bibliography{deed}

\appendix

\noindent This is the appendix for the main \ours paper.
Here we discuss the architecture and results of LaTr++, and the results of our reproduced OFA \vs the original OFA results.

\section{LaTr++}
\label{sec_supp:latr}
\begin{figure}[t!]
    \centering
    \includegraphics[width=\linewidth]{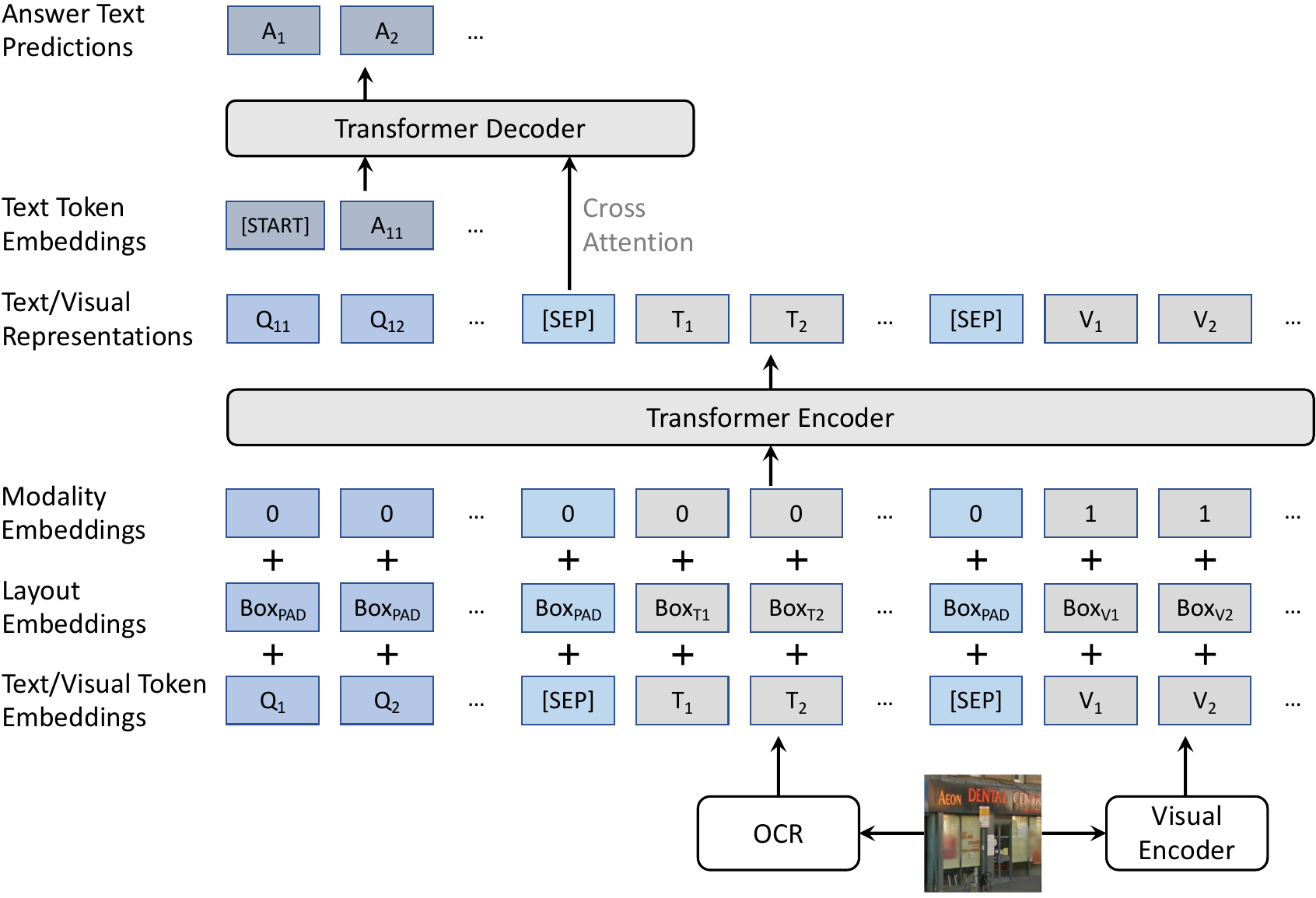}
    \caption{The architecture of LaTr++ for text-VQA.}
    \label{fig_supp:latr}
\end{figure}

LaTr \cite{biten2022latr} obtains the state-of-the-art results on the text-based visual question answering (text-VQA) task. 
LaTr uses multi-modal encoder-decoder transformer models which takes OCR text, layout, and visual features as inputs.
We improve LaTr by replacing the ViT-based vision backbone \cite{dosovitskiy2020image} with simple multi-layer perceptrons and adding more unsupervised pre-training tasks, following DocFormerv2 \cite{appalaraju2023docformerv2}.
We refer to the improved LaTr as LaTr++.
See Figure \ref{fig_supp:latr} for the architecture of LaTr++ and more details below.

\subsection{Architecture}

In LaTr++, given an input image, we resize the image to size 500x384 and split the image into 196 patches with patch size 32x32.
Instead of using ViT \cite{dosovitskiy2020image}, we simply use a linear projection layer to generate 196 visual token embeddings for each patch.
We further use one more linear layer with the intention of compressing the extracted 196 visual tokens to only 128 visual tokens.
These visual tokens are then concatenated with word embeddings, from here the architecture is identical to LaTr \cite{biten2022latr} and T5 \cite{raffel2020exploring}.
Arguably our LaTr++ architecture is much more simpler than LaTr \cite{biten2022latr} as we do not have a pre-trained ViT as a dependency, hence our model has less number of parameters for equal model size compared to LaTr \cite{biten2022latr}.

\begin{table*}[t]
    \centering
    \small
    \begin{tabular}{l | c c c}
        \toprule
         & ST-VQA (ANLS) $\uparrow$ & TextVQA (Accuracy) $\uparrow$ & OCR-VQA (Accuracy) $\uparrow$ \\
        \midrule
        LaTr$_{\text{base}}$ & 68.3 & 59.5 & 67.5 \\
        LaTr++$_{\text{base}}$ & \textbf{69.7} & \textbf{61.1} & \textbf{68.4} \\
        \midrule
        \midrule
        LaTr$_{\text{large}}$ & 70.2 & 61.1 & - \\
        LaTr++$_{\text{large}}$ & \textbf{70.3} & \textbf{63.1} & \textbf{70.1}\\
        \bottomrule
        
    \end{tabular}
    \caption{Accuracy comparison between LaTr++ and LaTr on ST-VQA, TextVQA, and OCR-VQA validation sets. The best results are in \textbf{bold face}.}
    \label{tab_supp:latr}
\end{table*}

\begin{table*}[t]
    \centering
    \small
    \begin{tabular}{l | c c | c c c | c c c | c c}
        \toprule
        \multirow{2}{*}{} & \multicolumn{2}{c|}{VQA} & \multicolumn{3}{c|}{RefCOCO} & \multicolumn{3}{c|}{RefCOCO+} & \multicolumn{2}{c}{RefCOCOg}\\
        \cline{2-11}
        & test-dev$\uparrow$ & \multicolumn{1}{c|}{test-std$\uparrow$} & val $\uparrow$ & testA $\uparrow$ & \multicolumn{1}{c|}{testB $\uparrow$} & val $\uparrow$ & testA $\uparrow$ & \multicolumn{1}{c|}{testB $\uparrow$} & val-u $\uparrow$ & test-u $\uparrow$ \\
        \midrule
        OFA (original) & 79.4 & 79.5 & 90.1 & 92.9 & 85.3 & 85.8 & 89.9 & 79.2 & 85.9 & 86.6 \\
        OFA (reproduced) & 79.3 & 79.4 & 90.6 & 92.5 & 85.9 & 85.7 & 89.9 & 78.6 & 87.2 & 87.6 \\
        \bottomrule
    \end{tabular}
    \caption{Accuracy comparisons between the original OFA and our reproduced OFA.}
    \label{tab_supp:ofa}
\end{table*}

\subsection{Pre-training}
\label{sec_sup:pretrain}

We use the IDL dataset\footnote{\url{https://www.industrydocuments.ucsf.edu/}} described in the main paper to pre-train the LaTr++ models.
We use the standard T5 denoising pre-training task \cite{raffel2020exploring} as in the original LaTr paper \cite{biten2022latr}. 
In addition, to make the LaTr++ a more competitive baseline we add two more unsupervised pre-training tasks at the encoder: a) Line prediction task - in order to teach the model the relative position semantic information between text tokens, we randomly pick two text tokens and ask the model to predict how many lines are between them. There are only three labels: 0, 1 and 2. Any text token pairs that have more than 2 lines between them are assigned to 2 because distant text tokens are not related and the model does not need the precise number of lines between them. b) Token-to-grid task - To utilize global information the task involves creating a virtual 3x3 grid and asking the network to predict which grid each text token falls in.
Losses of all three tasks, \ie, standard denoising, line prediction, and token-to-grid, are added to form the final pre-training loss for LaTr++.

\subsection{Settings}
\label{sec_supp:settings}

We evaluate on four text-VQA datasets: DocVQA \cite{mathew2021docvqa}, OCR-VQA \cite{mishra2019ocr}, ST-VQA \cite{biten2019scene}, and TextVQA \cite{singh2019towards}.
DocVQA is a VQA dataset dedicated to document text understanding, and OCR-VQA focuses on question-answering on book covers.
ST-VQA and TextVQA contain natural images of everyday scenes with textual information and require the understanding of the text in the image to answer the question. Following \cite{biten2022latr}, we use Amazon Textract\footnote{\url{https://aws.amazon.com/textract/}} for DocVQA, Amazon Text-in-Image\footnote{\url{https://docs.aws.amazon.com/rekognition/latest/dg/text-detecting-text-procedure.html}} for ST-VQA and TextVQA, and Rosetta~\cite{borisyuk2018rosetta} for OCR-VQA, to extract text information from images.

\subsection{Implementation Details}
\label{sec_supp:implement_detail}

We pre-train our models on the Industrial Document Library (IDL) dataset\footnote{\url{https://www.industrydocuments.ucsf.edu/}}, using the tasks described in Section \ref{sec_sup:pretrain}.
We add deep-supervision loss of the T5 denoising task on all decoder layers for \oursnospace, because we found it considerably improves the generative ability of shallow layers, as discussed in our main paper.  We pre-train the base version with deep supervision on 5M IDL data for 30 epochs. For the large variation, to reduce the computational costs while achieving competitive performance, we firstly pre-train the model on 64M IDL data for 1.5 epochs without deep supervision, and then pre-train it on 64M IDL data with deep supervision using batch size 18 for 60k steps. The models are then fine-tuned on each dataset following the same settings as in~\cite{powalski2021going,biten2022latr}. We follow the convention of fine-tuning on the combination of ST-VQA and TextVQA training sets when evaluating on these two datasets \cite{biten2022latr}. The confidence score threshold \th is selected using cross-validation, specifically, 0.99 on DocVQA and 0.95 for other three datasets.

\subsection{Results}

Here we compare LaTr++ to LaTr on three text-VQA datasets: ST-VQA \cite{biten2019scene}, TextVQA \cite{singh2019towards}, and OCR-VQA \cite{mishra2019ocr}.
For ST-VQA and TextVQA, we train LaTr++ on the combination of ST-VQA and TextVQA training sets, following LaTr \cite{biten2022latr}.
For OCR-VQA, we train LaTr++ on the OCR-VQA training set only.
All results are reported on the validation sets of these three datasets.
As we can see in Table \ref{tab_supp:latr}, LaTr++ obtains better results than state-of-the-art approach LaTr on the text-VQA task.

\subsection{Analyses on the Number of Parameters}

\begin{table}[t]
    \centering
    \small
    \begin{tabular}{l | c c c}
        \toprule
        & \multicolumn{3}{c}{\# Parameters}\\
        & Baseline & Ours & Previous \\
        \midrule
        LaTr++ (-b) & 232M & 239M & 503M \\
        LaTr++ (-L) & 750M & 774M & 1507M \\
        \bottomrule
    \end{tabular}
    \caption{\# parameters of the baseline model (Baseline), our approach (Ours), and previous approaches (Previous).}
    \label{tab:parameters}
\end{table}

Compared to the previous multi-exit models, one advantage of our multi-exit model is fewer number of parameters. 
For LaTr++, the base version (-b) has 232M parameters with hidden size ($d_{model}$) 768, 12 encoder layers, and 12 decoder layers. The large version (-L) has 750M parameters with $d_{model}$ 1024, 24 encoder layers, and 24 decoder layers. Each adaptation module consists of a linear layer ($d_{model}\times d_{model}$ parameters) followed by layer normalization ($d_{model}$ parameters). Therefore, each adaption module only has 0.6M parameters (-b) and 1.05M parameters (-L). The adaption modules from all layers only increase $\sim$3\% of the full model parameters. In contrast, a generation head has 24.7M parameters (-b) and 32.9M parameters (-L) due to the output dimension (32128). Using unshared generation heads increases the total number of parameters by $>$100\%. So our adaptation module has much fewer parameters than unshared generation heads in previous approaches.
See Table~\ref{tab:parameters} for more details.

\section{OFA Results}

We use the original pre-trained OFA model and the same fine-tuning procedure as in \cite{wang2022ofa} to reproduce the OFA results and train DEED, using the official code\footnote{\url{https://github.com/OFA- Sys/OFA}}.
The only exception is RefCOCOg - we fine-tune our model on top of the RefCOCO fine-tuned model, because RefCOCOg has fewer training samples than RefCOCO and RefCOCO+, which makes the accuracy on RefCOCOg inferior if we fine-tune our model on RefCOCOg directly.
Here we compare our reproduced OFA results and the original OFA results on VQAv2~\cite{balanced_vqa_v2} for VQA and RefCOCO/RefCOCO+/RefCOCOg \cite{yu2016modeling,mao2016generation} for referring expression comprehension.
As we can see, our reproduced results are close to the reported numbers.

\end{document}